\documentclass[sigconf]{acmart} 
\AtBeginDocument{%
  }

\setcopyright{acmlicensed}
\copyrightyear{2025}
\acmYear{2025}

\acmDOI{XXXXXXX.XXXXXXX}
\acmConference[ACM MM 2025]{33rd ACM International Conference on Multimedia}{October 27--31,
  2025}{Dublin, Ireland}
\acmISBN{978-1-4503-XXXX-X/2018/06}

\acmSubmissionID{2}
\usepackage{multirow}
\usepackage{tabularx}
\usepackage{url}
\usepackage{booktabs}
\usepackage{listings}
\newcommand{\newpara}[1]{\vspace{8pt}\noindent\textbf{#1}}
\usepackage{siunitx}

\copyrightyear{2025}
\acmYear{2025}
\setcopyright{acmlicensed}\acmConference[MM '25]{Proceedings of the 33rd ACM International Conference on Multimedia}{October 27--31, 2025}{Dublin, Ireland}
\acmBooktitle{Proceedings of the 33rd ACM International Conference on Multimedia (MM '25), October 27--31, 2025, Dublin, Ireland}
\acmDOI{10.1145/3746027.3761981}
\acmISBN{979-8-4007-2035-2/2025/10}

\begin{document}

\title{Pindrop it! Audio and Visual Deepfake Countermeasures for Robust Detection and Fine-Grained Localization}

\author{Nicholas Klein}
\email{nklein@pindrop.com}
\affiliation{%
  \institution{Pindrop Security Inc.}
  \city{Atlanta}
  \state{Georgia}
  \country{USA}
}

\author{Hemlata Tak}
\email{hemlata.tak@pindrop.com}
\affiliation{%
  \institution{Pindrop Security Inc.}
  \city{Atlanta}
  \state{Georgia}
  \country{USA}
}

\author{James Fullwood}
\email{james.fullwood.i@pindrop.com}
\affiliation{%
  \institution{Pindrop Security Inc.}
  \city{Atlanta}
  \state{Georgia}
  \country{USA}
}

\author{Krishna Regmi}
\email{krishna.regmi@pindrop.com}
\affiliation{%
  \institution{Pindrop Security Inc.}
  \city{Atlanta}
  \state{Georgia}
  \country{USA}
}

\author{Leonidas Spinoulas}
\email{leonidas.spinoulas@pindrop.com}
\affiliation{%
  \institution{Pindrop Security Inc.}
  \city{Atlanta}
  \state{Georgia}
  \country{USA}
}

\author{Ganesh Sivaraman}
\email{gsivaraman@pindrop.com}
\affiliation{%
  \institution{Pindrop Security Inc.}
  \city{Atlanta}
  \state{Georgia}
  \country{USA}
}

\author{Tianxiang Chen}
\email{tchen@pindrop.com}
\affiliation{%
  \institution{Pindrop Security Inc.}
  \city{Atlanta}
  \state{Georgia}
  \country{USA}
}

\author{Elie Khoury}
\email{ekhoury@pindrop.com}
\affiliation{%
  \institution{Pindrop Security Inc.}
  \city{Atlanta}
  \state{Georgia}
  \country{USA}
}

\renewcommand{\shortauthors}{Nicholas Klein et al.}


\begin{abstract}
The field of visual and audio generation is burgeoning with state-of-the-art methods. This proliferation of new techniques underscores the need for robust solutions for detecting synthetic content in videos. In particular, when fine-grained alterations via localized manipulations are performed in visual, audio, or both domains, these subtle modifications add challenges to the detection algorithms.
This paper presents solutions for the problems of deepfake video classification and localization. The methods were submitted to the 2025 ACM Multimedia 1M-Deepfakes Detection  Challenge, achieving the best performance in the temporal localization task and a top four ranking in the classification task for the \emph{TestA} split of the evaluation dataset. 

\end{abstract}




\maketitle

\section{Introduction}
With rapid advancements in generative techniques, the creation of synthetic videos
has become affordable, fast, and easily accessible to the public.
While these technologies are used in domains such as AI-generated movies~\cite{Fakemoviegeneration} and gaming~\cite{Fakevideogame}, they also pose a serious threat when exploited by malicious actors for disinformation, financial fraud~\cite{financialfrauds}, or hiring scams~\cite{hiringscams}.
Deepfake content refers to manipulated media that can contain modifications in the audio or visual domains and can span over a localized region or over the full duration of the media. Visual manipulation of the face region in particular poses privacy and security risks. There exists a multitude of generative engines for performing face-swap, where the face in the video is replaced with another face, and face-reenactment, where facial attributes are modified to achieve synchronization with target spoken words. Similarly, there are a large number of publicly available text to speech and voice conversion models, many of which enable the synthesis of public figures’ voices or the cloning of any voice for which just a minute of example speech is available. The rapid development of these models which are capable of hyperrealistic manipulations highlights the urgent need for robust deepfake detection systems.

To foster progress in audio-visual deepfake detection, the 1M-Deepfakes Detection Challenge~\cite{cai2024av} played an important role in initiating research in this domain. As part of the challenge, the organizers released the AV-Deepfake1M dataset in 2024 and the AV-Deepfake1M++ extended and enhanced version in 2025. This new dataset comprises over 2M samples across thousands of speakers, making it one of the most comprehensive datasets for multi-modal deepfake detection. It introduces audio manipulations through word-level deletions, insertions, and replacements, followed by fine-grained alignment of lip movements and facial expressions to match the altered speech content~\cite{cai2024av}. 
Based on the training and validation data labels, the audio manipulations are done using YourTTS~\cite{casanova2022yourtts} and VITS~\cite{kim2021vits} engines, and the visual manipulations consist of the face reenactment methods Diff2Lip~\cite{mukhopadhyay2023diff2lip} and TalkLip~\cite{wang2023seeing} which particularly focus on lip synchronization. Furthermore, the dataset contains localized modifications, with each video having very few words altered in the audio and/or visual streams.
To address this, we propose an ensemble of specialized networks that independently target audio and visual manipulations.
For each of the classification and localization tasks, we propose specific architectures, and each model is optimized for its respective task, as detailed in later sections. Our proposed approaches have shown strong performance in the
\textbf{2025 ACM MM 1M-Deepfakes Detection Challenge}, performing competitively in the detection task and achieving the top performance in the localization task. 

The main contributions of our work are as follows: 
we adapt existing audio and visual countermeasures for the task of partial deepfake detection; we explore the novel composition of existing audio and visual countermeasure backbones with the localization training paradigm of ActionFormer~\cite{zhang2022actionformer}, resulting in a first place localization performance on the \emph{TestA} set~\cite{cai2025av}.



\section{Related Work}
\textbf{Audio Deepfakes:}
Previous studies focused mainly on utterance-level detection by designing models that capture global artifacts introduced during speech synthesis or voice conversion. For instance, state-of-the-art SSL-AASIST~\cite{tak2022automatic} proposed a heterogeneous graph attention network to learn both temporal and spectral representations, while ASDG~\cite{xie2023domain} employed domain generalization through aggregation and separation to enhance robustness across unseen conditions. Although these models perform on fully spoofed utterances, their performance drops significantly when detecting partial deepfakes, where only specific segments are manipulated. This task, known as Partial Spoofed Detection (PSD)~\cite{zhang2021initial} requires fine-grained temporal resolution to capture subtle modifications. 

To address PSD, recent studies focused on models operating at varying temporal resolutions.
~\cite{wu2022partially,zhang2022partialspoof,cai2023waveform,li2023multi,cai2024integrating,xie2024efficient}.~\cite{wu2022partially} introduced a question-answering-inspired framework with self-attention mechanisms to detect partially fake audios.~\cite{zhu2023local} adopted a hybrid multi-instance learning with local self-attention to learn temporal dependencies. 
Furthermore, works such as~\cite{zhang2022partialspoof,li2023multi} used self-supervised learning (SSL)-based front-end features with multi-resolution detection heads for improving accuracy and generalization. 
Other approaches~\cite{cai2023waveform,cai2024integrating} improved frame-level localization using SSL-based backbones with specialized transform blocks to predict frame-wise boundary offset probabilities.~\cite{zhong2024enhancing} further advanced performance by introducing a Boundary-aware Attention Mechanism (BAM), combining boundary enhancement and frame-wise attention modules to better distinguish real and fake segments at the frame level.

\textbf{Visual Deepfakes: }
Early work in video deepfake detection trained a fully temporal convolution network (FTCN)~\cite{zhang2022exploring} that reduced the spatial dimension to 1, followed by a temporal Transformer head. This led to suppressing the spatial artifacts in the video frames, limiting the model's generalization capability. STIL~\cite{gu2021spatiotemporal} proposed spatial and temporal inconsistency modules to learn spatiotemporal differences over adjacent frames in both horizontal and vertical directions. Recent NACO work~\cite{zhang2024learning} learns natural consistency representations of real face videos in a self-supervised manner for generalizable deepfake detection on videos.
Some visual deepfake countermeasures focus on identifying artifacts in the mouth movements. LIPINC~\cite{datta2024exposing} focuses on local and global inconsistencies in the mouth region. LipForensics~\cite{haliassos2021lips} leverages the mouth movement information from the embeddings of a frozen VSR encoder, training a multi-scale temporal convolutional network (MS-TCN) backend to perform deepfake detection.

\textbf{Audio Visual Deepfakes:} 
Video deepfake localization has recently gained significant attention, aiming to accurately detect manipulated temporal segments by leveraging multi-modal (audio-visual) representation. Several works have explored the integration of cross-modal feature learning using advanced temporal localization decoders~\cite{lin2019bmn,su2021bsn++,zhang2022actionformer}. LAV-DF~\cite{cai2022you} was the first work to introduce a content-driven multi-modal audio-visual dataset and proposed the BA-TFD framework, which combines contrastive learning, frame-level classification, and the boundary matching network (BMN)~\cite{lin2019bmn}. BA-TFD+~\cite{cai2023glitch} further improved performance by introducing a multi-scale transformer and a BSN++~\cite{su2021bsn++}-based boundary detection module. BA-TFD and BA-TFD++ later served as baselines in the 1M AV deepfake detection challenge 2025.

Audio-visual temporal forgery detection (AV-TFD) model~\cite{liu2023audio} also utilized BMN as its localization backbone and proposed a cross-modal attention mechanism, alongside embedding-level fusion, for robust and generalized audio-visual representation learning. AVH-Align~\cite{Smeu2025CircumventingSI} looked at the temporal alignment between audio and visual content to identify inconsistencies in fake segments, however this can fail when the fake segments have good audio-visual alignment even when detectable artifacts in the audio and visual streams are present.
Recently, UMMAFormer model~\cite{zhang2023ummaformer} enhanced localization accuracy by introducing a temporal abnormality attention module and a parallel cross-attention feature pyramid network combined with an ActionFormer-based decoder~\cite{zhang2022actionformer} for localization. Audio-visual models can fail in cases where one of the modalities may not be present in the data, e.g., silent videos, and thus learning of modality-specific models and their fusion is proposed in this work.

\section{Methodology}


\subsection{Deepfake Classification (Task 1)}

\begin{figure}[!t] 
    \centering
    \includegraphics[width=0.9\linewidth]{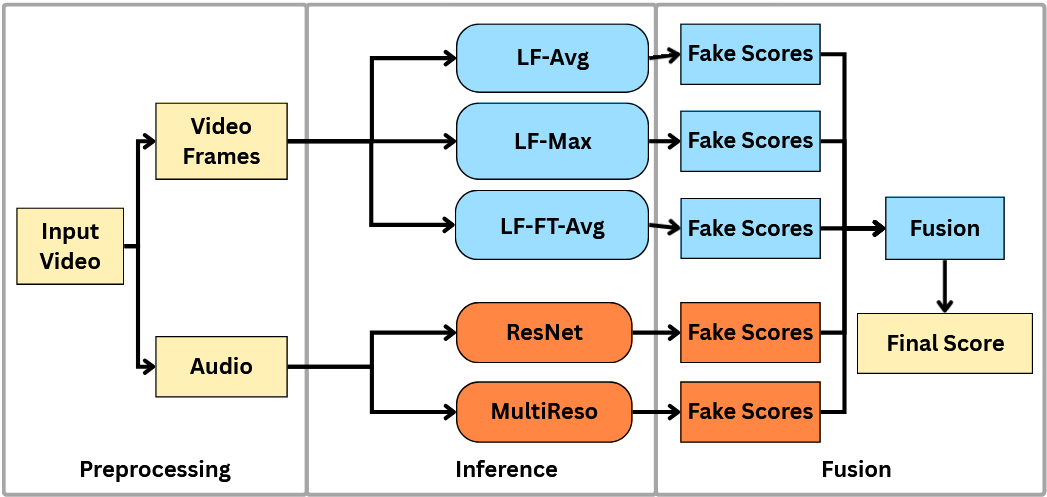} 
    \caption{Task~1 overview.}
    \label{fig:task1_overview}
    \Description[Task~1 overview]{Overview of our proposed approach for Task~1.}
    \vspace{-0.5cm}
\end{figure}

\begin{figure*}[!t] 
    \centering
    \includegraphics[width=0.9\linewidth]{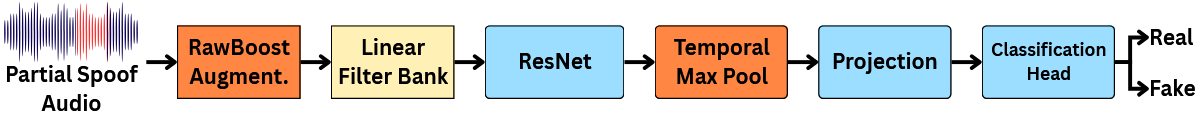} 
    \caption{Proposed ResNet-based architecture for deepfake classification task.}
    \label{fig:task1_audio1_model}
    \Description[Task~1 ResNet architecture]{Proposed ResNet-based architecture for deepfake classification task.}
\end{figure*}

For a given video, the task is to predict a score corresponding to the likelihood that the video contains \textit{any} synthetic content. This score can be used downstream for classification by thresholding. A video is considered real if both the audio and visual components are fully real, and it is expected to receive a low fake score. A video should receive a high fake score if any part of the video is synthetic. This means that a video is considered fake if either its audio or visual components include any synthetic content.
Video-level labels (real/fake) are available for the training data. However, the specific timestamps of fake segments within the videos are not. Figure~\ref{fig:task1_overview} provides an overview of the architecture of the system for deepfake detection, each block being discussed in detail in the following subsections.


\subsubsection{Audio Models}
We propose a variety of audio models that focus on discriminating partially fake speech from real speech.\\

\noindent \textbf{ResNet:}
To enhance the diversity and variability of the training data, we apply RawBoost~\cite{tak2022rawboost} data augmentation\footnote{\url{github.com/TakHemlata/RawBoost-antispoofing}}, a widely adopted technique in audio deepfake detection. 
RawBoost introduces nuisance variability by adding linear and non-linear convolutive noise as well as impulsive signal-dependent additive noise directly at the raw waveform level. We use the same rawboost configuration parameters as in the original work~\cite{tak2022rawboost}. For front-end feature extraction, we employ an 80-dimensional log linear filter bank (LFB) from 20\si{s} audio segments, using a window length of 20\si{ms} with a 10\si{ms} frame shift. The audio waveforms are either truncated or zero-padded to ensure a fixed 20\si{s} input.  

For the backbone model, we utilize a deep residual network (ResNet)~\cite{he2016deep} to learn higher-level representations from low-level acoustic features. In particular, we optimize the channel capacity and depth of the network by adopting ResNet-152 to improve performance. Our implementation builds on the ResNet architecture from the Wespeaker toolkit\footnote{\url{https://github.com/wenet-e2e/wespeaker/tree/master/wespeaker/models}} with a slight modification in the pooling strategies. Instead of average-pooling or attentive statistical-pooling, we apply a temporal max-pooling layer across the frame axis to capture the most discriminative temporal frame. A fully-connected layer then extracts 256-dimensional embeddings from the pooled features, which are finally passed through a classification head to determine whether the input speech is real or fake. Our proposed architecture is illustrated in Figure~\ref{fig:task1_audio1_model}.


\begin{figure}[!t]
    \centering
    \includegraphics[width=0.75\linewidth]{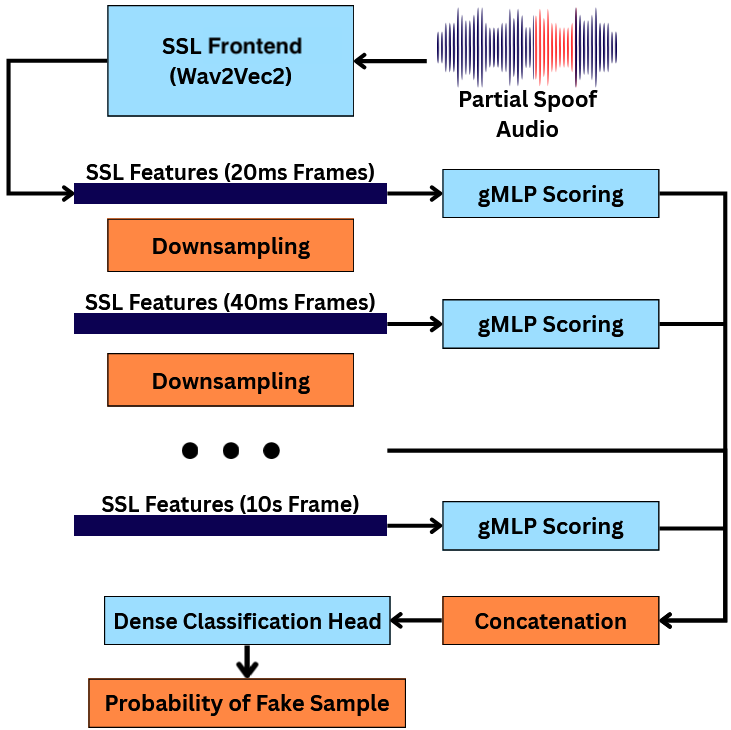}
    \caption{Multi-Resolution gMLP pyramid with Wav2Vec SSL for full file classification.}
    \label{fig:task1_audio2}
    \Description[Task~1 Multi-Resolution architecture]{Multi-Resolution gMLP pyramid with Wav2Vec SSL for full file classification.}
    \vspace{-10pt}
\end{figure}

\textbf{MultiReso gMLP:}
For full utterance classification, we draw on the multi-resolution countermeasure proposed by~\cite{multireso}, which combines a pretrained SSL frontend (in this case, a Wav2Vec2 transformer~\cite{wav2vec2}) and a pyramidal downsampling structure, which produces combinations of the source features at successively coarser temporal resolutions. The architecture is shown in Figure~\ref{fig:task1_audio2}. Each temporal resolution has a separate gMLP-scoring module that produces a sequence of logits for each feature frame at that resolution. The logits for each resolution are concatenated and then passed to a standard dense classifier head, which is trained on the utterance-level audio labels. The primary improvement here over previous works is the use of all scales of features in the final classification rather than only using the final lowest resolution scale features. The classifier operates on 10\si{s} audio chunks, with the final reported score being the maximum chunk score for all chunks of the given audio file. This ensures that a long file with fake segments only in a single chunk will still be classified as fake.

\subsubsection{\label{sec:task1_video_models}Visual models}
We leverage the LipForensics model introduced by Haliassos et. al~\cite{haliassos2021lips} to detect synthetic visuals in the mouth region. As described in their work, videos are processed to yield mouth crops before being encoded by a frozen visual speech recognition (VSR) (aka “lipreading”) frontend. The backend multi-scale temporal convolutional network (MS-TCN) is trained on the extracted frontend features. To facilitate training for partial deepfake detection using video-level labels only, we train and predict on full-length videos as opposed to 25-frame clips as the original authors did. Furthermore, we train three variations of this model.

\textbf{LF-avg}: The MS-TCN backend is trained from scratch and the temporal pooling layer uses an average operation.

\textbf{LF-max}: The MS-TCN backend is trained from scratch and the temporal pooling layer uses a max operation.

\textbf{LF-ft-avg}: The MS-TCN backend is initialized with the publicly available pretrained weights\footnote{\href{https://github.com/ahaliassos/LipForensics}{https://github.com/ahaliassos/LipForensics}} provided by~\cite{haliassos2021lips}, which have been learned by training for deepfake detection on various publicly available datasets. Notably, the aforementioned pretraining is not performed on partial deepfake data. We then fine-tune the pretrained MS-TCN backend on the partial-deepfake data of the challenge, where the temporal pooling layer uses an average operation.

\subsubsection{Fusion}
The score outputs of the two audio models and three visual models are fused using score-level polynomial logistic regression fusion. 
We perform z-score normalization of the model scores.
We then compute the 2nd order polynomial terms of the scores and the logistic regression coefficients for the polynomial terms using a grid search of the regularization parameter on the validation data.
An empirical study on the validation data determined that this outperformed a traditional linear logistic regression fusion approach.

\subsection{\label{sec:Deepfake Localization}Deepfake Localization (Task 2)}
\begin{figure}[!t] 
    \centering
    \includegraphics[trim={0cm 0.05cm 0cm 0cm},clip,width=1\linewidth]{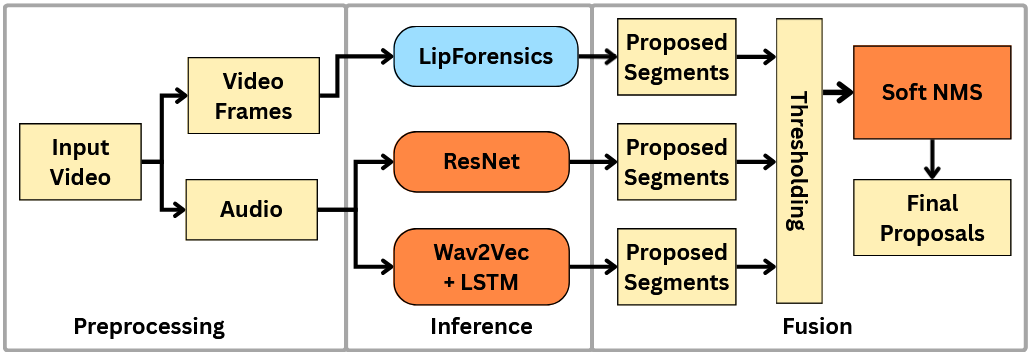} 
    \caption{Task~2 overview.}
    \label{fig:task2_overview}
    \Description[Task~2 overview.]{Overview of our proposed approach for Task~2.}
    \vspace{-0.4cm}
\end{figure}



For a given video, the task is to predict which segments of the video are synthetic, if any. A video may contain any number of fake segments, each of which should be identified. The fake segments to be identified may contain synthetic audio, synthetic video, or both. Each predicted fake segment of the video is composed of the start and end timestamps of the segment and a fake score corresponding to the likelihood that the segment is fake. The specific timestamps of the fake segments within the videos are available for the training data. We train three models, two audio models and one visual model, to address the localization task. In the remainder of this section, we describe the details of our localization training paradigm that are common across our three models. The fine-grained localization of short fake segments within the audio or visual recordings of the video is very challenging and requires a precise detection approach. To address this, we employ a localization training paradigm inspired by ActionFormer~\cite{zhang2022actionformer}, which utilizes a frame-wise classification head alongside a dedicated segment boundary regression head.

For each individual audio and visual model, a backbone network is utilized to learn features from the audio or visual data. The backbones utilized for the detection task are leveraged here with an adaptation: the output of the backbone network maintains the temporal dimension to enable frame-wise classification and regression for localization training. Each frame-level feature is then processed by two separate heads: a classification head and a regression head. Both heads consist of two 256-dimension fully connected layers with ReLU activations. The classification head predicts the likelihood of each frame being fake, and the regression head predicts the temporal offsets (in seconds) from the center of each frame to the start and end of the fake segment that it falls within. The classification and regression heads are trained jointly. For the classification task, Focal Loss~\cite{lin2017focal} with a parameter value of $\alpha$ = 0.9 is used to focus on the harder samples and to automatically deal with the severe class imbalance in the training data (“fake” frames are much less common than “real” frames). For the boundary regression task, Distance-IoU Loss~\cite{zheng2020distance} is used, and the loss is computed only on frames that fall within a fake segment according to the ground truth labels. The total loss is then computed as a weighted sum of the classification and regression losses, where the regression component is scaled by a coefficient of 0.03 to account for its larger scale. During inference, every analysis frame will have a predicted fake score, start offset, and end offset.
Figure~\ref{fig:task2_overview} illustrates the overall system architecture designed for Task~2, with details in the following subsections.



\subsubsection{Audio Models}
Similar to Task~1, we also used two different audio models for the temporal localization task.

\begin{figure}[!t] 
    \centering
    \includegraphics[trim={0cm 0.1cm 0cm 0cm},clip,width=1.0\linewidth]{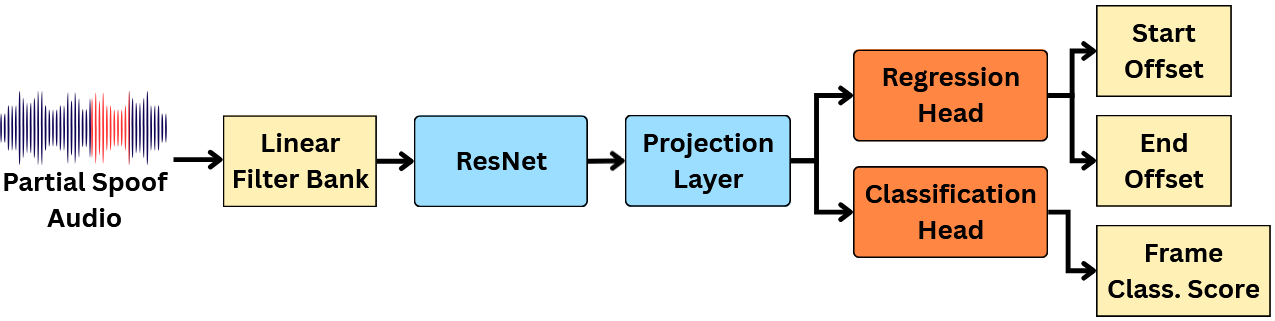} 
    \caption{ResNet-based end-to-end pipeline for frame-level fake speech detection and localization task.}
    \label{fig:task2_audio1_model}
    \Description[Task~2 ResNet pipeline]{ResNet-based end-to-end pipeline for frame-level fake speech detection and localization task.}
    \vspace{-10pt}
\end{figure}

\newpara{ResNet}: To perform frame-level localization of fake segments, we adapt the ResNet-152 backbone that we utilized in the detection task to be used in composition with the localization training paradigm described in section~\ref{sec:Deepfake Localization}. Specifically, the temporal max-pooling layer is omitted from the ResNet-152 architecture to preserve the temporal resolution of the intermediate feature representations. Our proposed end-to-end pipeline for detecting and temporally localizing deepfake speech at the frame level is illustrated in Figure~\ref{fig:task2_audio1_model}. For task~2, audio was segmented into 20\si{s} windows with a frame resolution of 40\si{ms} and LFB acoustic features were extracted accordingly. To maintain consistent input length across the samples, the final audio segment was zero-padded, and padded frames were masked out during loss computation to prevent distortion in training. Each frame-level feature output from the ResNet-152 backbone was passed through both the classification and regression heads as described in section~\ref{sec:Deepfake Localization}. ReLU activation was applied to the outputs of the regression head to ensure non-negative start and end offset predictions.

\begin{figure}[!t]
    \centering
    \includegraphics[width=0.8\linewidth]{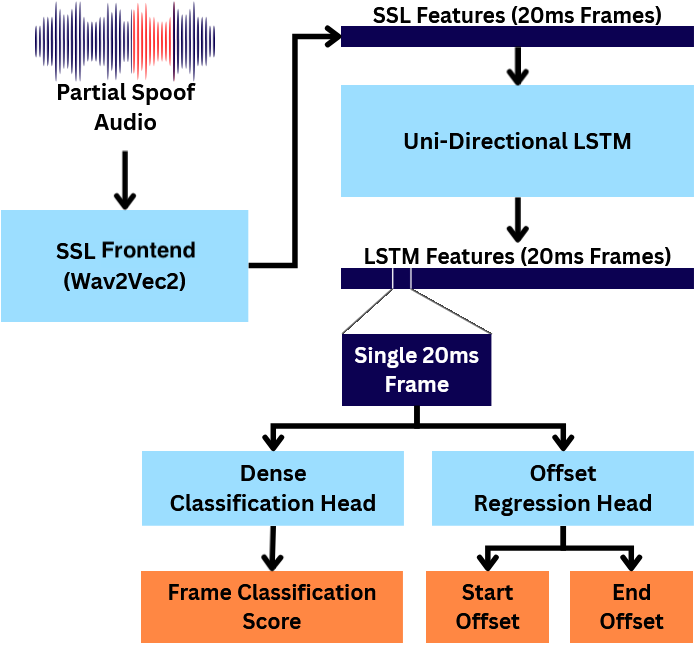}
    \caption{Wav2Vec-based SSL pipeline with LSTM for fine-grained frame-level detection and localization task. 
    }
    \label{fig:task2_audio2}
    \Description[Task~2 SSL pipeline]{Wav2Vec-based SSL pipeline with LSTM for fine-grained frame-level detection and localization task.}
    \vspace{-0.5cm}
\end{figure}

\newpara{SSL+LSTM}: Our second audio localization model employs the same frame-wise classification and regression strategy as the other localization models, but leverages the embedding capabilities of the Wav2Vec2~\cite{wav2vec2} transformer.
The block diagram, shown in Figure~\ref{fig:task2_audio2}, produces a series of audio frame embeddings at a rate of 50 frames per second (each frame encompassing 20\si{ms} of audio). Although these embeddings already contain some cross-frame information from the internal attention mechanisms of the transformer, we additionally incorporate a single-layer, unidirectional LSTM to learn additional temporal relationships between frames and provide an opportunity for class-specific duration modeling to occur. Without the LSTM layer, frame-wise classification tends to result in relatively noisy sequences, with single-frame misclassifications interrupting otherwise homogeneous regions. The output of the LSTM is a new sequence of 20\si{ms} frames.
The classification and regression heads follow the standard structure for this task, but replace the usage of \lstinline{torch.clamp} with the Softplus function to constrain the regression output to be strictly positive. This avoids the gradient discontinuities associated with simply clamping the regression output to the desired range.

\subsubsection{Visual Model}
The LipForensics model described in section~\ref{sec:task1_video_models} is adapted to be leveraged for localization. The preprocessing and VSR feature extraction process remains the same. However, the MS-TCN backend is modified by omitting the temporal pooling layer and replacing the video-level classification head by frame-wise classification head and adding a regression head. The model is trained as described in section~\ref{sec:Deepfake Localization}, and the outputs of the regression head are constrained with a ReLU activation function to enforce positive offsets. The MS-TCN backend is trained from scratch.

\subsubsection{Fusion}
The fake segments predicted for a given video from the ResNet, SSL+LSTM, and LipForensics models are combined into a single set of reduced predicted fake segments using Soft-NMS~\cite{bodla2017soft}. 
First, predicted segments with a fake score below 0.2 are first filtered out. Next, Soft-NMS is applied across the predicted segments of all algorithms jointly (instead of separately for audio and video algorithms). For all experiments, a standard deviation parameter of 0.8 was used when performing Soft-NMS. Notably, while a sigmoid activation function is applied to constrain the classification scores from both the SSL+LSTM model and the LipForensics model to a range of $(0, 1)$, the same constraint is not applied to the classification logits of the ResNet model. This allows the segment proposals of the ResNet model to take precedence over overlapping proposals from the other two models during Soft-NMS when the ResNet model has high confidence that the segment is fake. We empirically found that this resulted to improved performance on the validation set compared to constraining all three models to the same range.


\section{Experiments}

\subsection{Dataset and Metrics}
The AV-Deepfake1M++ dataset~\cite{cai2025av} is used for the challenge. It is a large-scale dataset with more than 2 million videos. The training and validation sets have 2,606 subjects shared between them. There are 4,503 subjects shared between the \emph{TestA} and \emph{TestB} subsets.
The total number of videos within each subset is shown in Table~\ref{tab:dataset_stats} along with the breakdown of samples that are real, fake audio real video (FARV), real audio fake video (RAFV), and fake audio fake video (FAFV). The videos have an audio sample rate of 16 KHz and a visual frame rate of 25 frames per second. The fake segments of the training and validation sets are very short, with an average duration of just 0.33\si{s}. Additional details of the composition and preparation of the dataset are discussed in~\cite{cai2025av}.




For Task~1, Area Under the Curve (AUC) is used as the evaluation metric. 
For Task~2, the average precision (AP) and average recall (AR) scores are used for evaluation. AP is computed at intersection over union (IoU) thresholds of 0.5, 0.75, 0.9, and 0.95. AR is computed at five values of N, the number of top proposals to consider: 50, 30, 20, 10, and 5. The set of IoU thresholds utilized to compute each value of AR@N is [0.5:0.95:0.05]. The overall localization performance metric is then computed as the weighted average of the four AP@IoU values and five AR@N values, giving equal weights 1/8 and 1/10 to the overall AP and AR, respectively.




\begin{table}[!t]
    \centering
    \caption{Number of samples in the AV-Deepfake1M++ dataset, categorized by real/fake class of audio and visual streams. Wherever `-', the values couldn't be computed as the labels were not shared. FARV: Fake Audio Real Video, RAFV: Real Audio Fake Video, and FAFV: Fake Audio Fake Video.}
    \vspace{-0.4cm}
    \label{tab:dataset_stats}
  \resizebox{0.5\textwidth}{!}{
\begin{tabular}{|c|c|c|c|c|c|}

\hline
\textbf{Subset}& \textbf{Real}&\textbf{FARV}&\textbf{RAFV} &\textbf{FAFV}  & \textbf{Total} \\
\hline
Train & 297,509 &	258,149&	261,759	&281,800	 & 1,099,217\\
\hline
Val & 20,226 &	18,299	&18,465	&20,336	   & 77,326\\
\hline
\emph{TestA}& 287,517& - & - & -  & 828,318 \\
\hline
\emph{TestB}  & 22,810 & - & - & - & 46,293\\
\hline
Overall& 317,735 & - & - & - & 2,051,154  \\
\hline
\end{tabular}}
\end{table}



\begin{table}[!t]
    \centering
    \caption{Task~1 performance in AUC (\%) on validation set. }
    \vspace{-0.3cm}
    \label{tab:task1_val_classification}
  \resizebox{0.5\textwidth}{!}{
\begin{tabular}{|c|c|c|c|}

\hline
\textbf{Method} & \textbf{Audio Labels}&\textbf{Visual Labels}&\textbf{AV Labels} \\
\hline
ResNet&98.44\%&-&85.67\%\\
\hline
MultiReso &97.61\%&-&81.61\%\\
\hline
LF-avg&-&99.22\%&81.77\%\\
\hline
LF-max&-&99.22\%&81.93\%\\
\hline
LF-ft-avg&-&99.08\%&82.34\%\\
\hline
Fused & 79.45\%&91.98\%&99.77\% \\
\hline
\end{tabular}}
\vspace{-15pt}

\end{table}





\begin{table*}[!t]
  \caption{Validation results analysis of audio and visual models on localization task. `audio-visual' label is fake if either modality is fake, otherwise real. 
  }
  \vspace{-0.4cm}
  \label{tab:localization_val}
  \Large
  \resizebox{1\textwidth}{!}{
  \begin{tabular}{|c|c|c|c|c|c|c|c|c|c|c|c|}
    \hline
    \textbf{Models} &  \textbf{Label} &  \textbf{Score} &	\textbf{AP@0.5}&	\textbf{AP@0.75}	&\textbf{AP@0.9}&	\textbf{AP@0.95}&	\textbf{AR@50}	&\textbf{AR@30}	&\textbf{AR@20}	&\textbf{AR@10}	&\textbf{AR@5} \\
    \hline
    ResNet &	audio &	87.76 & 89.62 &	88.11 &	85.64 &	81.28 &	89.41 &	89.41 &	89.41 &	89.38 &	89.22   \\
    \hline
    SSL+LSTM &	audio &	74.00 &	81.58 &	66.11 &	49.39 &	40.67 &	89.84 &	89.71 &	89.50 &	88.41 &	85.35  \\
    \hline
    LipForensics  &	visual &	79.49 &	97.86 &	90.57 &	55.59 &	22.54 &	93.49 &	93.41 &	93.25 &	92.06 &	89.43  \\
    \hline
     &	audio &	91.66 &	93.05 &	91.28 &	87.38 &	81.83 &	95.77 &	95.55 &	95.24 &	94.52 &	93.58  \\
    Fusion &	visual &	73.41 &	65.54 &	61.72 &	47.05 &	32.39 &	96.32 &	96.10 &	95.75 &	95.75 &	92.94  \\
     &	audio-visual &	89.17 &	95.20 &	91.91 &	81.22 &	67.83 &	95.34 &	95.12 &	94.79 &	93.83 &	92.35  \\
    \hline
  \end{tabular}
  }
  \vspace{-5pt}
\end{table*}

\begin{table}[!t]
    \centering
    \caption{Task~1 performance comparison with baseline systems and top-5 teams~\cite{cai2025av} on \emph{TestA} set.}
    \vspace{-0.3cm}
    \label{tab:classification}

\begin{tabular}{|c|c|c|}

\hline
\textbf{Method/Team} & \textbf{AUC (\%)} \\
\hline
Baseline Xception\cite{chollet2017xception} &  55.09 \\
\hline
Mizhi Labs & 91.78 \\
\hline
\textbf{Pindrop Labs (Ours)} & \textbf{92.49} \\
\hline
KLASS & 92.78 \\
\hline
WHU\_SPEECH & 93.07 \\
\hline
XJTU SunFlower Lab & 97.83 \\
\hline
\end{tabular}
\vspace{-5pt}
\end{table}

\begin{table}[!t]
    \centering
    \caption{Task~2 performance comparison with baseline systems and top-5 teams~\cite{cai2025av} on \emph{TestA} set.}
    \vspace{-0.3cm}
    \label{tab:localization_comparsion}
    \Large
\resizebox{0.475\textwidth}{!}{
\begin{tabular}{|c|c|c|c|c|}

\hline
\textbf{Method/Team} & \textbf{Score (\%)}&\textbf{Avg. AP (\%)}&\textbf{Avg. AR (\%)}\\
\hline
Baseline BA-TFD\cite{cai2023glitch} &13.54  &2.78&24.29\\
\hline
Baseline BA-TFD++\cite{cai2022you} &14.71  &4.10&25.31\\
\hline
KLASS &35.36  &28.13&42.59\\
\hline
WHU{\_}SPEECH &41.20 &25.41 &57.16\\
\hline
Purdue-M2  &50.87  &46.76&55.48\\
\hline
Mizhi Labs &55.00  &44.81&65.19\\
\hline
\textbf{Pindrop Labs (Ours)} & \textbf{67.20}&55.85&78.55 \\
\hline
\end{tabular}}
\vspace{-15pt}
\end{table}



\subsection{Analysis on the Validation Set}

We analyze validation results as the test-set labels has not been released.

\textbf{Partial deepfake detection.} The results of each individual Task~1 model assessed against its corresponding domain labels (audio or visual) and overall labels are shown in Table~\ref{tab:task1_val_classification}. Between the two audio models, we observe that the ResNet-based model was slightly more successful in classifying partially fake audios compared to the MultiReso model, achieving 98.44\% and 97.61\% AUC, respectively. 
On the video labels, all three visual models achieve an AUC above 99\%, suggesting that the visual deepfake generation methods may be easier to detect than the audio generation methods. 
We experimented with using embeddings from a pretrained CLIP model~\cite{yan2024df40} rather than VSR embeddings. This gave inferior results, which we do not present here. The use of VSR embeddings was particularly effective on this data because the visual synthetic generation methods are lip-sync methods. Additionally, the VSR features are exposed to temporal context when extracting the frame embeddings, whereas CLIP-based features are not. Although the individual visual models performed similarly, using all three for fusion proved beneficial. On the overall audio-visual labels, the fused model achieved an AUC of 99.77\%.

\textbf{Deepfake localization.} The localization metrics of each Task~2 model along with the fusion are shown in Table~\ref{tab:localization_val}. Among the three individual models assessed on their corresponding domains' labels, the ResNet model achieves the highest score of 87.76. The key differentiator that makes this model the strongest is its AP metrics, especially at higher IoU thresholds, highlighting its ability to accurately predict segment boundaries.
For the LipForensics model, we observe slightly higher AR metric values, around four points higher than the audio models for $N\geq20$. This observation is in line with our finding from Task~1 that the artifacts of the visual generation methods used in this dataset are easier to detect than those of the audio generation methods. However, when observing the AP metrics of the LipForensics model, we find significantly lower scores of 55.59 and 22.54 at the higher IoU thresholds of 0.9 and 0.95, respectively. This highlights that the weakness of the LipForensics model is in accurately predicting the segment boundaries. This may be explained by the significantly lower temporal resolution of the visual stream: at only 25 FPS, the 40\si{ms} gaps in the video frames are 12\% of the length of the average duration fake segment, 0.33\si{s}.

When fusing the three models, the complementarity of their detection capabilities led to an increase in the AR and AP metrics at the more lenient IoU thresholds of 0.5 and 0.75. For the more aggressive IoU thresholds of 0.9 and 0.95, AP reduces because the LipForensics model is not as accurate in predicting segment boundaries. However, the reduction is moderate compared to the LipForensics model's AP metrics because of our fusion strategy, which enables the more accurate boundary predictions of the ResNet model to take precedence over overlapping proposals from the other two models when it is confident.
\subsection{\emph{TestA} Results}


Table~\ref{tab:classification} and Table~\ref{tab:localization_comparsion} present a performance comparison between our proposed system, the top four competing teams, and the baselines reported in~\cite{cai2025av}. On Task~1, our system ranked 4th with an AUC score of 92.49\%, comparable to the 3rd place team. On Task~2, our system significantly outperformed the other teams, demonstrating the robustness of our approach for the challenging temporal localization task. With a score of 67.20, our system surpassed the second best method by an absolute score of 12.20 and achieved higher AP and AR values at all evaluated thresholds.
These are slightly lower than the performance on the validation set, as shown in Table~\ref{tab:localization_val} (last row). This is likely because the subjects and manipulation techniques overlap between the train and validation sets but not so with the \emph{TestA} set.
\section{Conclusions}

In this work, we present our approach to the 2025 ACM Multimedia 1M-Deepfakes Detection  Challenge. For the first task of partial deepfake detection, we train a ResNet-based model and a multi-resolution SSL-based model for detecting partial deepfake audio, as well as three variants of a LipForensics-based model for detecting partial deepfake visuals. Through polynomial score-level fusion of these five models, we achieve a fourth-place AUC score of 92.49\% on the \emph{TestA} set. For the second task of deepfake localization, we explore the novel composition of the Resnet, SSL + LSTM, and LipForensics backbones with the temporal localization training paradigm of Actionformer~\cite{zhang2022actionformer}, achieving the first-place localization score of 67.20 on the \emph{TestA} set. 

Regarding future work, further exploration of the fusion strategy for the localization task could be beneficial. In particular, methods that explicitly learn to fuse the proposals of individual models may outperform the simple application of soft-NMS. Additionally, methods of learning from audio and visual information together can be explored as a way to improve performance over single-modality fused systems. Finally, additional data augmentation techniques could be used during the training to improve generalization of the proposed approaches to the test sets.
\newpage



\bibliographystyle{unsrt}
\bibliography{sections/refs}


\end{document}